\documentclass[runningheads]{llncs}

\usepackage[mobile]{eccv}

\usepackage{eccvabbrv}
\usepackage{graphicx}
\usepackage{colortbl}  
\usepackage{xcolor}
\usepackage{booktabs}
\usepackage{comment}
\usepackage{amssymb,mathrsfs,amsmath}
\usepackage[colorlinks,linkcolor=blue]{hyperref}
\usepackage{marvosym}
\usepackage[accsupp]{axessibility}  

\usepackage{hyperref}

\usepackage{orcidlink}

\begin{document}

\title{Infinite-ID: Identity-preserved Personalization via ID-semantics Decoupling Paradigm}

\titlerunning{Infinite-ID}

\author{Yi Wu\inst{1}$^{\star}$ \and
Ziqiang Li\inst{1}\thanks{First two authors contributed equally to this work.} 
\and
Heliang Zheng\inst{1}\and
Chaoyue Wang\inst{2} \and
Bin Li\inst{1}\thanks{Corresponding Author.}
}
%
%
\institute{University of Science and Technology of China, China \and
The University of Sydney, Australia
} 
\maketitle
\centerline {\href{https://infinite-id.github.io/}{https://infinite-id.github.io/}}

\begin{abstract}

Drawing on recent advancements in diffusion models for text-to-image generation, identity-preserved personalization has made significant progress in accurately capturing specific identities with just a single reference image. However, existing methods primarily integrate reference images within the text embedding space, leading to a complex entanglement of image and text information, which poses challenges for preserving both identity fidelity and semantic consistency. To tackle this challenge, we propose \textbf{Infinite-ID}, an ID-semantics decoupling paradigm for identity-preserved personalization. Specifically, we introduce identity-enhanced training, incorporating an additional image cross-attention module to capture sufficient ID information while \textit{deactivating the original text cross-attention module} of the diffusion model. This ensures that the image stream faithfully represents the identity provided by the reference image while mitigating interference from textual input. Additionally, we introduce a feature interaction mechanism that combines a mixed attention module with an AdaIN-mean operation to seamlessly merge the two streams. This mechanism not only enhances the fidelity of identity and semantic consistency but also enables convenient control over the styles of the generated images. Extensive experimental results on both raw photo generation and style image generation demonstrate the superior performance of our proposed method.

\end{abstract}

\begin{figure}[!tbp]
    \centering
    \includegraphics[scale=0.18]{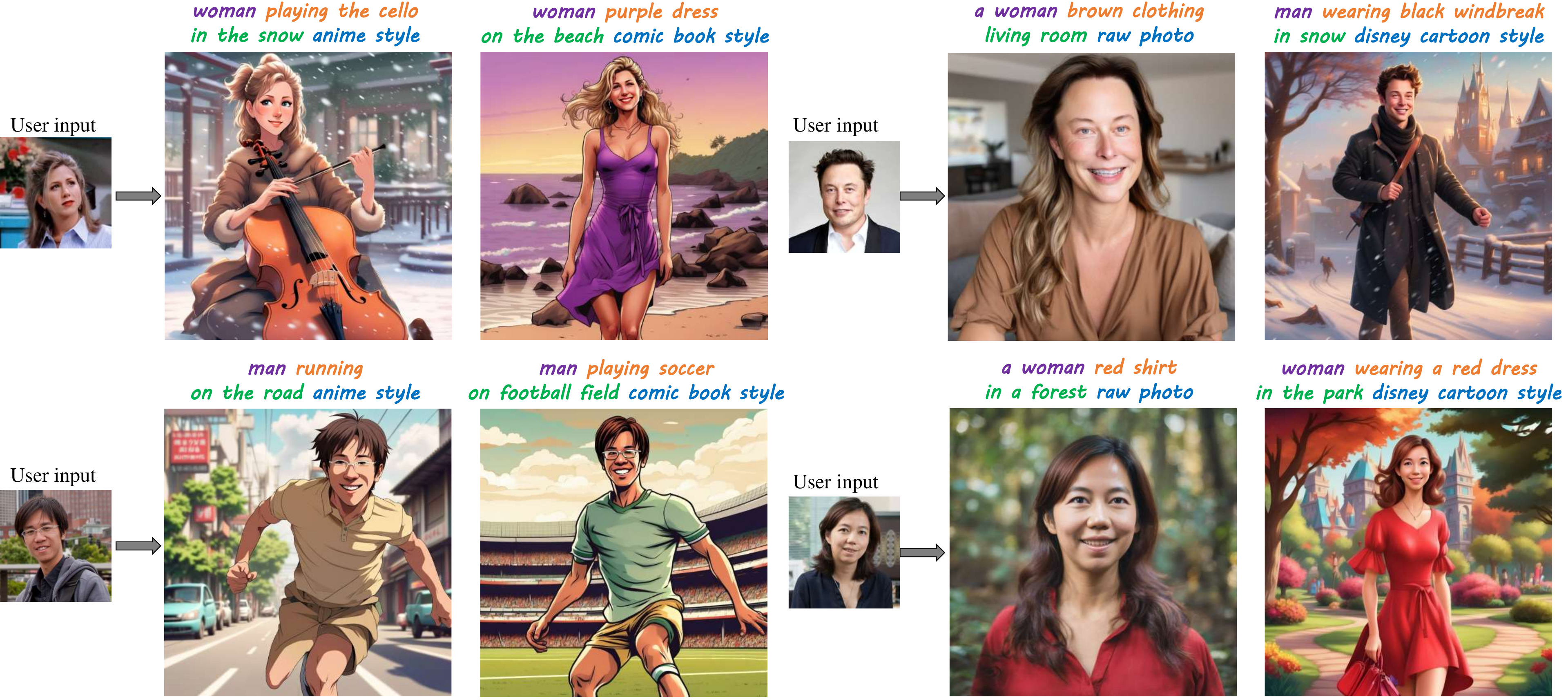}
    \caption{
With just a single reference image, our Infinite-ID framework excels in synthesizing high-quality images while maintaining superior identity fidelity and text semantic consistency in various styles. 
}
    \label{image-firstphoto}
\end{figure}

\section{Introduction}
Human photo synthesis \cite{li2022fakeclr,wu2024domain} has experienced notable advancements, particularly with the introduction of large text-to-image diffusion models such as Stable Diffusion (SD) \cite{rombach2022high}, Imagen \cite{saharia2022photorealistic}, and DALL-E 3 \cite{betker2023improving}. Benefit from personalized text-to-image generation, recent researches focus on \textit{Identity-preserved personalization}. This specialized area aims to produce highly customized photos that faithfully reflect a specific identity in novel scenes, actions, and styles, drawing inspiration from one or more reference images. This task has garnered considerable attention, leading to the development of numerous applications, including personalized AI portraits and virtual try-on scenarios. In the context of Identity-preserved personalization, the emphasis is placed on maintaining the invariance of human facial identity (ID), requiring a heightened level of detail and fidelity compared to more general styles or objects.

Recent tuning-free methods exhibit promise for large-scale deployment, yet they face a notable challenge in balancing the trade-off between the fidelity of identity representation (ID fidelity) and the consistency of semantic understanding conveyed by the text prompt. This challenge arises due to the inherent entanglement of image and text information. Typically, tuning-free methods extract ID information from reference images and integrate it into the semantic information in two distinct ways. The first type, exemplified by PhotoMaker \cite{li2023photomaker}, incorporates text information with ID details in the text embedding space of the text encoder. While this merging approach aids in achieving semantic consistency, it compresses image features into the text embedding space, thereby weakening the ID information of the image and compromising identity fidelity. The second type, demonstrated by IP-Adapter \cite{ye2023ip}, directly injects ID information into the U-Net of the diffusion model through an additional trainable cross-attention module. Although this approach aims to enhance the strength of ID information for improved fidelity, it tends to favor the image branch during training, consequently weakening the text branch and compromising semantic consistency. In summary, existing methods entangle image and text information, resulting in a significant trade-off between ID fidelity and semantic consistency (as illustrated in Fig. \ref{image-scatter}).

To address the entanglement between image and text information, we propose \textbf{Infinite-ID}, an innovative approach to personalized text-to-image generation. Our method tackles the trade-off between maintaining high fidelity of identity and ensuring semantic consistency of the text prompt by implementing the ID-semantics decoupling paradigm. 
Specifically, we adopt identity-enhanced training that introduces an additional image cross-attention module to capture sufficient ID information and deactivate the original text cross-attention module to avoid text interference during training stage. Accordingly, our method can faithfully capture identity information from reference image, significantly improving the ID fidelity. 
Additionally, we employ a novel feature interaction mechanism that leverages a mixed attention module and an AdaIN-mean operation to effectively merge text information and identity information. Notably, our feature interaction mechanism not only preserves both identity and semantic details effectively but also enables convenient control over the styles of the generated images (as depicted in Fig. \ref{image-firstphoto}).
Our contributions are summarized as:
\begin{itemize}
\item We propose a novel ID-semantics decoupling paradigm to resolve the entanglement between image and text information, acquiring a remarkable balance between ID fidelity and semantic consistency in \textit{Identity-preserved personalization}.

\item We propose a novel feature interaction mechanism incorporating a mixed attention module and an AdaIN-mean operation to effectively merge ID information and text information and also conveniently control the styles of the generated image in diffusion models.
\item Experimental results demonstrate the excellent performance of our proposed method as compared with current state-of-the-art methods on both raw photo generation and style image generation.
\end{itemize}

\section{Related Works}

\subsection{Text-to-image Diffusion Models}

Text-to-image diffusion models, such as those explored in \cite{rombach2022high,saharia2022photorealistic,betker2023improving,zhang2023adding,zhang2023text,kawar2023imagic}, have garnered significant attention due to their impressive image generation capabilities. Current research endeavors aim to further enhance these models along multiple fronts, including the utilization of high-quality and large-scale datasets \cite{schuhmann2021laion,schuhmann2022laion}, refinements to foundational architectures \cite{rombach2022high,saharia2022photorealistic,betker2023improving}, and advancements in controllability \cite{zhang2023adding,inoue2023layoutdm,ruiz2023dreambooth}. Present iterations of text-to-image diffusion models typically follow a two-step process: first, encoding the text prompt using pre-trained text encoders such as CLIP \cite{radford2021learning} or T5 \cite{raffel2020exploring}, and then utilizing the resulting text embedding as a condition for generating corresponding images through the diffusion process. Notably, the widely adopted Stable Diffusion model \cite{rombach2022high} distinguishes itself by executing the diffusion process in latent space instead of the original pixel space, leading to significant reductions in computation and time costs. An important extension to this framework is the Stable Diffusion XL (SDXL) \cite{podell2023sdxl}, which enhances performance by scaling up the U-Net architecture and introducing an additional text encoder. Thus, our proposed method builds upon the SDXL. However, our method can also be extended to other text-to-image diffusion models.


\subsection{Identity-preserved personalization}

Identity-preserved personalization aims to generate highly customized photos that accurately reflect a specific identity across various scenes, actions, and styles, drawing inspiration from one or more reference images. Initially, \textit{tuning-based methods}, exemplified by DreamBooth \cite{ruiz2023dreambooth} and Textual Inversion \cite{gal2022image}, employ images of the same identity (ID) to fine-tune the model. While these methods yield results with high fidelity in preserving facial identity (ID), a significant drawback emerges: the customization of each ID necessitates a time investment of 10-30 minutes \cite{kumari2023multi}, consuming substantial computing resources and time. This limitation poses a significant obstacle to large-scale deployment in commercial applications. Consequently, recent advancements in \textit{tuning-free methods} \cite{xiao2023fastcomposer,ye2023ip,li2023photomaker,gal2023encoder,wei2023elite,li2023blip,chen2023anydoor,ma2023subject,chen2023photoverse} have been introduced to streamline the generation process. These methods specifically leverage the construction of a vast amount of domain-specific data and the training of an encoder or hyper-network to represent input ID images as embeddings or LoRA weights within the model. Post-training, users need only input an image of the ID for customization, enabling personalized generation within seconds during the inference phase. These tuning-free methods typically contain two  distinct manners.

On one hand, methods \cite{li2023photomaker,xiao2023fastcomposer,peng2023portraitbooth,achlioptas2023stellar} incorporate text information alongside identity details within the text embedding space of the text encoder. For example, PhotoMaker \cite{li2023photomaker} extracts identity embeddings from single or multiple reference images and merges them with corresponding class embeddings (e.g., "man" and "woman") in the text embedding space of the text encoder. While this stacking operation aids in achieving semantic consistency, it compresses image features into the text embedding space, leading to compromised identity fidelity. On the other hand, some studies \cite{ye2023ip,chen2023photoverse,wei2023elite} directly integrate identity information into the U-Net of the diffusion model. IP-Adapter \cite{ye2023ip} distinguishes itself by incorporating a additional cross-attention layer for each existing cross-attention layer within the original UNet model. This approach merges identity information with semantic details directly within the U-net but leads to distortion of the semantic space of the U-Net model. Consequently, this compromises the semantic consistency of the text prompt. 



In summary, existing methods entangle the identity and text information, leading to a significant trade-off between ID fidelity and semantic consistency. To mitigate these limitations, we propose an identity-enhanced training to capture ID and text information separately. Moreover, we design an effective feature interaction mechanism leveraging a mixed attention module and an AdaIN-mean operation to preserve both identity and semantic details while also enabling convenient control over the styles of generated images.

\subsection{Attention Control in Diffusion model}

Previous studies have investigated various attention control techniques within diffusion models. Hertz \textit{et al.} \cite{hertz2023style} employed a shared attention mechanism, concatenating and applying an AdaIN module on the key and value between reference and synthesis images within the self-attention layer to ensure style-consistent image generation using a reference style. Cao \textit{et al.} \cite{cao2023masactrl} utilized a mutual self-attention approach to achieve consistent image generation and non-rigid image editing, wherein the key and value of the synthesis image were replaced with those of the reference image within the self-attention layers of the diffusion model. Similarly, Shi \textit{et al.} \cite{shi2023zero123++} proposed a method termed reference attention, enabling consistent multi-view generation of target objects by concatenating the key and value features between the condition signal and the synthesis image in the self-attention layers. Wang \textit{et al.} \cite{xiao2023fastcomposer} and Avrahami \textit{et al.} \cite{avrahami2023break} exploited attention maps within the cross-attention layers to guide the optimization process towards disentangling learned concepts in personalized generation tasks.

\begin{figure}[!tbp]
    \centering
    \includegraphics[scale=0.28]{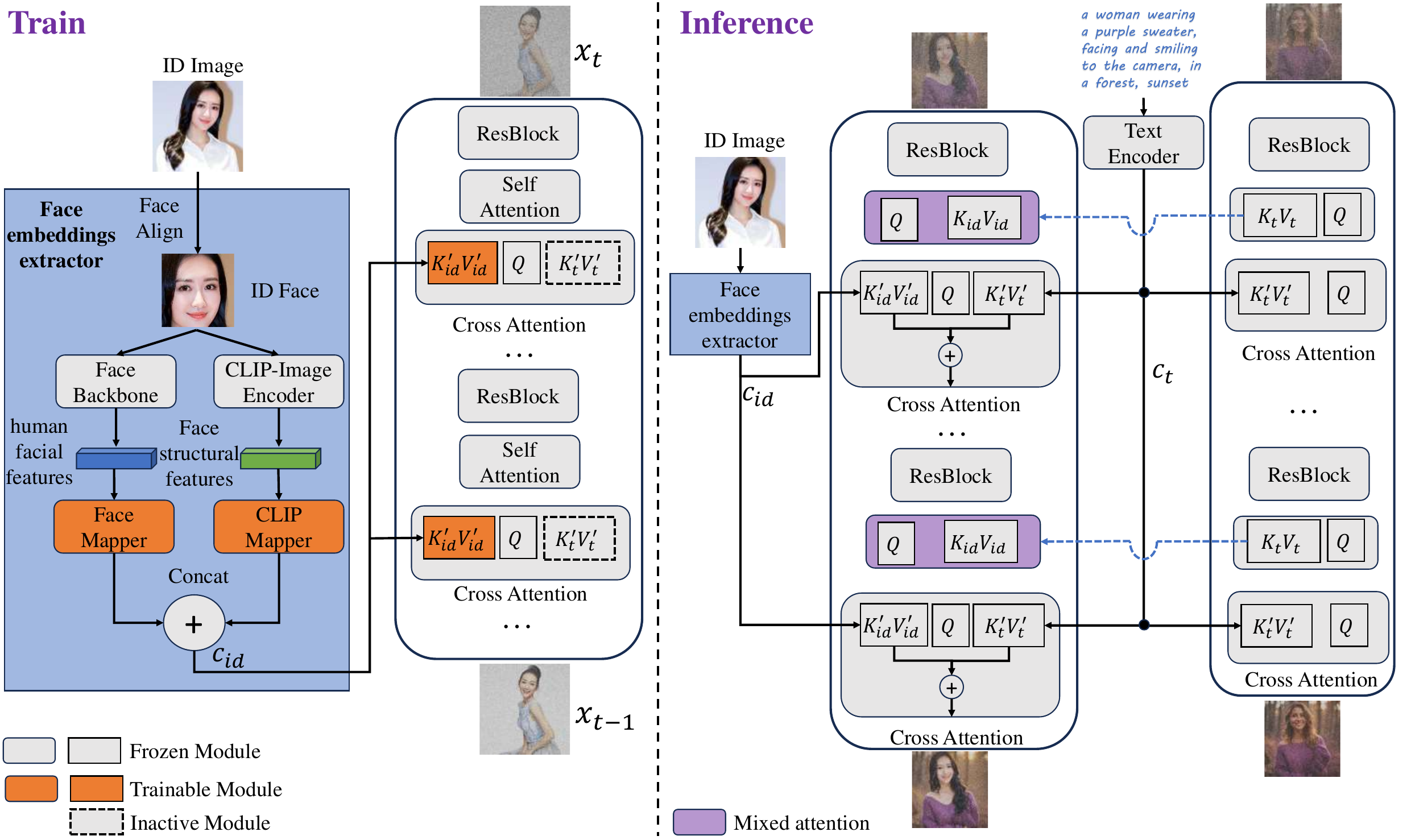}
    \caption{\textbf{Framework of ID-semantics Decoupling Paradigm}. In the training phase, we adopt Face embeddings extractor to extract rich identity information and identity-enhanced training for faithfully representing the identity provided by the reference image while mitigating interference from textual input. In the inference stage, a mixed attention module is introduced to replace the original self-attention mechanism within the denoising U-Net model, facilitating the fusion of both identity and text information.}
    \label{image-decoupling}
\end{figure}

\section{Method}

\subsection{Preliminaries}

\noindent\textbf{Stable Diffusion XL.}
Our method builds upon Stable Diffusion XL \cite{podell2023sdxl}, comprising three core components: a Variational AutoEncoder (VAE) denoted as $\xi(\cdot)$, a conditional U-Net \cite{ronneberger2015u} represented by $\epsilon_\theta(\cdot)$, and two pre-trained text encoders \cite{Radford_2021} denoted as $\Theta_1(\cdot)$ and $\Theta_2(\cdot)$. Specifically, given a training image $x_0$ and its corresponding text prompt $T$, the VAE encoder $\xi$ transforms $x_0$ from its original space $R^{H\times W\times 3}$ to a compressed latent representation $z_0 = \xi (x_0)$, where $z_0 \in R^{h\times w \times c}$ and $c$ denotes the latent dimension. Subsequently, the diffusion process operates within this compressed latent space to conserve computational resources and memory. Once the two text encoders process the text prompt $T$ into a text embedding $c=\text{Concat}(\Theta_1(T), \Theta_2(T))$, the conditional U-Net $\epsilon_\theta$ predicts the noise $\epsilon$ based on the current timestep $t$, the $t-$th latent representation $z_t$, and the text embedding $c$. The training objective is formulated as follows:

\begin{equation}
    L_\text{diffusion}=E_{z_t,t,c,\epsilon\in N(0,I)}[||\epsilon-\epsilon_\theta(z_t,t,c)||_{2}^{2}].\\
\label{equa-diffusion}
\end{equation}

\noindent\textbf{Attention mechanism in diffusion models.}
The fundamental unit of the stable diffusion model comprises a resblock, a self-attention layer, and a cross-attention layer. The attention mechanism is represented as follows:
\begin{equation}
    \text{Attn}(Q,K,V)=\text{Softmax}(\frac{QK^T}{\sqrt{d}})V,
\end{equation}
where $Q$ denotes the query feature projected from the spatial features generated by the preceding resblock, $K$ and $V$ represent the key and value features projected from the same spatial features as the query feature (in self-attention) or the text embedding extracted from the text prompt (in cross-attention).

\subsection{Methodology}

\begin{figure}[!tbp]
    \centering
    \includegraphics[scale=0.65]{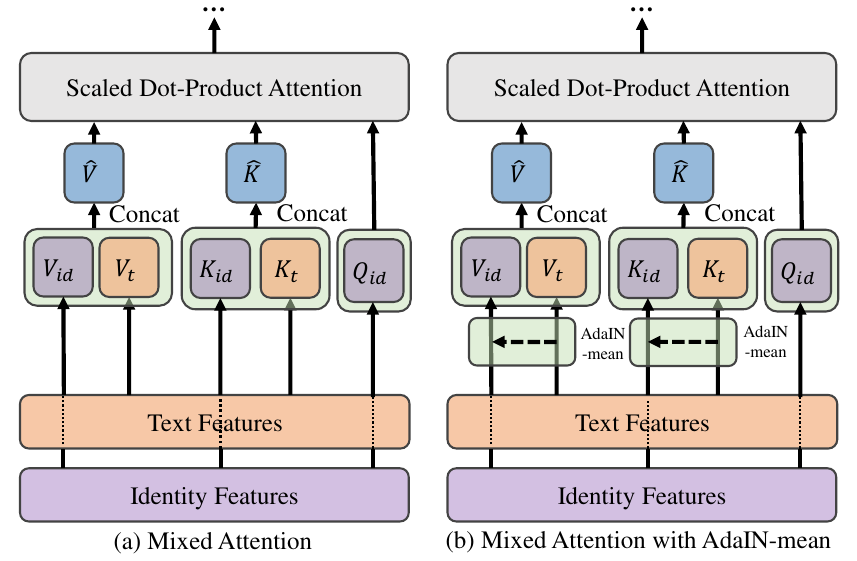}
    \caption{\textbf{Mixed attention mechanism}. On the left side, we employ mixed attention to fuse identity and text information. This involves concatenating their respective key and value features and subsequently applying mixed attention, where identity features are updated based on the concatenated key and value features. On the right side, for style merging, we introduce an additional AdaIN-mean operation (as depicted in Eq. \ref{equa-AdaINMean}) to the concatenated key and value features.}
    \label{image-MixedAttention}
\end{figure}

\noindent\textbf{Overview.}
In this section, we introduce our ID-semantics decoupling paradigm, as illustrated in Fig. \ref{image-decoupling}, which effectively addresses the severe trade-off between high-fidelity identity and semantic consistency within identity-preserved personalization. Subsequently, we present our mixed attention mechanism, depicted in Fig. \ref{image-MixedAttention}, designed to seamlessly integrate ID information and semantic information within the diffusion model during the inference stage. Additionally, we utilize an adaptive mean normalization (AdaIN-mean) operation to precisely align the style of the synthesized image with the desired style prompts.


\noindent\textbf{ID-semantics Decoupling Paradigm.}
To faithfully capture high-fidelity identity, we implement a novel identity-enhanced strategy during the training stage, as depicted in Fig. \ref{image-decoupling}. Diverging from conventional methods \cite{chen2023photoverse,li2023photomaker,ye2023ip} that utilize text-image pairs for training, we opt to exclude the text prompt input and deactivate cross-attention modules for text embeddings within the U-Net model. Instead, we establish a training pair consisting of an ID image, where the face is aligned to extract identity information, and denoising image ($x_t$ in Fig. \ref{image-decoupling}). Both the denoising image used for training and the ID image belong to the same individual, but they vary in factors such as viewpoints and facial expressions. This approach fosters a more comprehensive learning process \cite{li2023photomaker}. We adopt Face embeddings extractor to accurately capture and leverage identity information from the input ID image.
Additionally, to seamlessly integrate identity information into the denoising U-Net model, we introduce an extra trainable cross-attention mechanism ($K'_{id}$ and $V'_{id}$ in Fig. \ref{image-decoupling}) for image embeddings.

Throughout the training phase, we exclusively optimize the parameters associated with the face mapper, CLIP mapper, and the image cross-attention module, while keeping the parameters of the pre-trained diffusion model fixed. The optimization loss closely resembles the original diffusion loss formulation (as delineated in Eq. \ref{equa-diffusion}), with the sole distinction being the shift from a text condition to an identity condition as the conditional input.
\begin{equation}
    L_\text{diffusion}=E_{z_t,t,c_{id},\epsilon\in N(0,I)}[||\epsilon-\epsilon_\theta(z_t,t,c_{id})||_{2}^{2}],\\
\end{equation}
where the $c_{id}$ is the identity embeddings of the input ID image.

During the inference phase, we align the desired face within the ID image to afford identity information. Following the approach adopted during the training phase, we utilize a face recognition backbone and a CLIP image encoder to extract identity features from the aligned face image. Leveraging the trained face mapper, clip mapper, and image cross-attention mechanisms, these identity features are seamlessly integrated into the denoising U-Net model. Subsequently, we compute the key and value features for self-attention in the original stable diffusion model, considering only the text prompt input. These text key and value features are instrumental in the mixed attention process (illustrated in Fig. \ref{image-MixedAttention}), facilitating the fusion of text and identity information. Moreover, to further augment the text information during the denoising process, we incorporate the original text cross-attention mechanism, integrating the resulting text hidden states with the output image hidden states obtained from the image cross-attention module.


\noindent\textbf{Face Embeddings Extractor.} 
We adopt a multifaceted approach by incorporating pre-trained models to extract facial features. Firstly, following the methodologies of prior research, we utilize a pre-trained CLIP image encoder as one of our facial feature extractors. Specifically, we leverage the local embeddings, comprising the last hidden states obtained from the CLIP image encoder, forming a sequence of embeddings with a length of N (N=257 in our implementation). Subsequently, we employ a CLIP mapper to project these image embeddings (with a dimensionality of 1664) to the same dimension as the text features in the pre-trained diffusion model. 
As elucidated in \cite{ye2023ip}, the features extracted by the CLIP image encoder are instrumental in capturing the structural information pertinent to the identity face within identity-preserved personalization tasks. Additionally, we leverage the backbone of a face recognition model as another facial feature extractor. As highlighted in \cite{ye2023ip}, features extracted by the face recognition backbone are adept at capturing the characteristics associated with  human facial features within identity-preserved personalization tasks. More specifically, we utilize the global image embedding derived from the extracted features and subsequently employ a face mapper to align the dimensionality (512 dimensions) of the extracted global image embedding with the dimensionality of the text features in the pre-trained diffusion model.

In summary, the identity embeddings $c_{id}$ corresponding to the ID image $x$ can be expressed as:

\begin{equation}
    c_{id}=\text{Concat}\bigg(\text{M}_\text{clip}\big(\text{E}_\text{clip}(\text{FA}(x))\big),\text{M}_\text{face}\big(\text{E}_\text{face}(\text{FA}(x))\big)\bigg),
\end{equation}
where $\text{Concat}(\cdot,\cdot)$, $\text{M}_\text{clip}(\cdot)$, $\text{E}_\text{clip}(\cdot)$, $\text{M}_\text{face}(\cdot)$, $\text{E}_\text{face}(\cdot)$, and $\text{FA}(\cdot)$ denote the concatenation function, CLIP mapper, CLIP image encoder, face mapper, face recognition backbone, and face alignment module\cite{zhang2016joint}, respectively.


\noindent\textbf{Mixed Attention Mechanism.}
As explored in previous studies \cite{wu2023tune,khachatryan2023text2video,cao2023masactrl}, the features in the self attention layers play a crucial role in consistency image generation (across-frame in text-to-video works), which indicates these features provide a refined and detailed semantics information. In our study, we extract features from the self-attention layers of the original text-to-image diffusion model to capture rich semantic information, represented as $K_{t}$ and $V_{t}$. We enhance the self-attention mechanism by incorporating it into a mixed attention framework, depicted in Fig. \ref{image-MixedAttention} (a). This fusion enables the integration of semantic features ($K_{t}$ and $V_{t}$) with the identity-based features ($K_{id}$ and $V_{id}$), thereby encapsulating identity information. Through this integration, the mixed attention mechanism seamlessly merges semantic details into the generated features across different resolutions. The formulation of the mixed attention mechanism is as follows:




\begin{equation}
\begin{aligned}
    &\text{Attn}_\text{mix}(Q,K,V)\triangleq\text{Attn}(Q,\hat{K},\hat{V})\\
    \text{w.r.t}\quad &\hat{K}=\text{Concat}(K_{id}, K_{t}),\quad \hat{V}=\text{Concat}(V_{id}, V_t),\\
    &K_{id}=W^{id}_k Z_{id},\quad K_{t}=W^{t}_k Z_{t},\\
    &V_{id}=W^{id}_v Z_{id},\quad V_{t}=W^{t}_v Z_{t},
\end{aligned}
\end{equation}
where $Z_{id}$ and $Z_t$ are the corresponding spatial features of generated features and semantic features, respectively. The parameters $W^{id}_k$, $W^{t}_k$, $W^{id}_v$, and $W^{t}_v$ correspond to the weights of the corresponding fully connected layers.

\noindent\textbf{Cross-attention Merging.}
To further refine semantic control, we incorporate text features into the identity feature within the cross-attention layers using the following formulation:
\begin{equation}
   \text{Attn}_\text{cross}(Q,K,V) \triangleq \text{Attn}(Q,K'_{id},V'_{id}) + \text{Attn}(Q,K'_t,V'_t),
\end{equation}
where $K'_{id}=\hat W^{id}_kc_{id}$, $V'_{id}=\hat W^{id}_vc_{id}$, $K'_t=\hat W^t_kc_t$, and $V'_t=\hat W^t_vc_t$. $c_{id}$ and $c_t$ are the identity embedding and text embedding, respectively. $\hat W^{id}_k$, $\hat W^{id}_v$, $\hat W^t_k$, and $\hat W^t_v$ correspond to the weights of the trainable fully connected
layers within cross-attention module.

\noindent\textbf{Style Information Merging.}
Inspired by \cite{hertz2023style}, we propose an adaptive mean normalization (AdaIN-mean) operation to further align the style of the synthesis image with the style prompts. Concretely, we align the key and value features projected from identity features in both mixed attention and cross-attention with the key and value features projected from text features, formulated as follows:
\begin{equation}
\begin{aligned}
{K}_{id}=\text{AdaIN-m}(K_{id}, K_t), \quad  {V}_{id}=\text{AdaIN-m}(V_{id}, V_t),\quad \text{For Mixed Attention }  \\
    K'_{id}=\text{AdaIN-m}(K'_{id}, K'_t), \quad  V'_{id}=\text{AdaIN-m}(V'_{id}, V'_t), \quad \text{For Cross Attention }
\end{aligned}
\end{equation}
where the AdaIN-mean operation ($\text{AdaIN-m}(\cdot)$) is defined as:
\begin{equation}
    \text{AdaIN-m}(x,y)=x-\mu(x)+\mu(y), 
    \label{equa-AdaINMean}
\end{equation}
where $\mu(x)\in R^{d_k}$ is the mean of key and value features across different pixels. The mixed attention with AdaIN-mean has been illustrated in Fig. \ref{image-MixedAttention} (b).
  
\section{Experiments}
After outlining the experimental setup in Sec. \ref{sec-exp_set}, we conduct a comparative analysis of raw photo generation and style image generation in Sec. \ref{sec-comparion}. Ablation studies, highlighting the significance of various components, are presented in Sec. \ref{sec:ablation}. Additionally, experiments involving multiple input ID images are detailed in \ref{sup-idmixed} of the supplementary materials. For further insights, Sec. \ref{sup-rawphoto} and Sec. \ref{sup-stylephoto} of the supplementary materials provide additional qualitative results for raw photo generation and style photo generation, respectively.

\subsection{Experimental Setup}
\label{sec-exp_set}
\noindent\textbf{Implementation Details.}
Our experiments leverage a pre-trained Stable Diffusion XL (SDXL) model \cite{podell2023sdxl}. For image encoding, we utilize the OpenCLIP ViT-H/14 \cite{zhai2019large} and the backbone of ArcFace \cite{Huang_Belongie_2017}. The SDXL model consists of 70 cross-attention layers, to each of which we append an additional image cross-attention module. Training is conducted on 16 A100 GPUs for 1 million steps, with a batch size of 4 per GPU. We employ the AdamW optimizer with a fixed learning rate of 1e-4 and weight decay set to 0.01. During inference, we employ the DDIM Sampler with 30 steps and guidance scale is set to 5.0. Training data are sourced from multiple datasets, including the LAION-2B dataset \cite{schuhmann2021laion}, the LAION-Face dataset \cite{zheng2022general}, and images collected from the internet. We curate a dataset where each individual is represented by multiple photographs.


\begin{figure}[!tbp]
    \centering
    \includegraphics[scale=0.55]{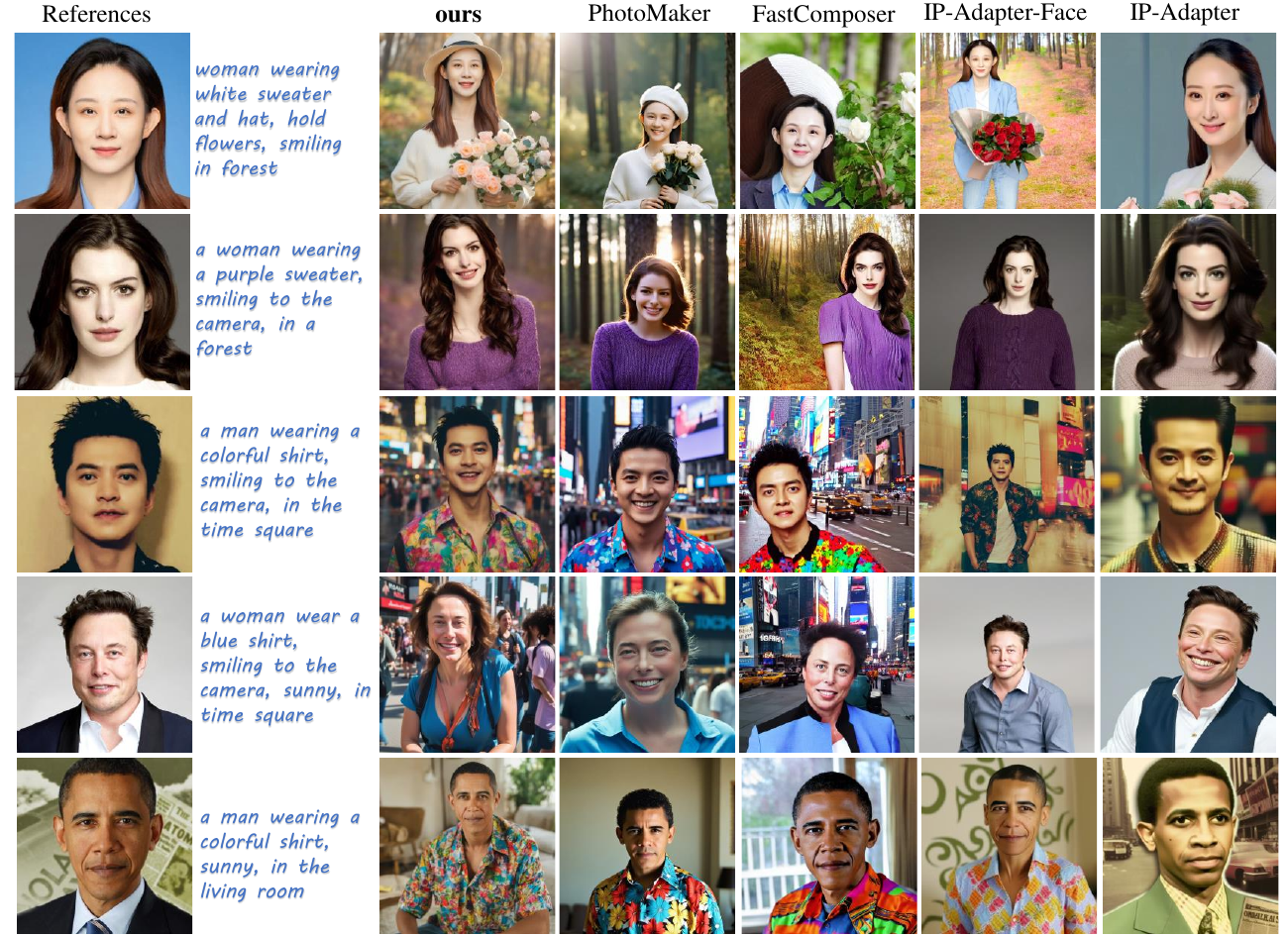}
    \caption{\textbf{Qualitative comparison on raw photo generation}. The results demonstrate that our Infinite-ID consistently maintains identity fidelity and achieves high-quality semantic consistency with just a single input image.}
    \label{image-realphoto-comparison}
\end{figure}

\noindent\textbf{Evaluation.}
We assess the efficacy of our approach in preserving both identity fidelity and semantic consistency. Specifically, we measure identity fidelity, utilizing metrics such as $M_\text{FaceNet}$ (measured by FaceNet \cite{schroff2015facenet}) and CLIP-I \cite{gal2022image}. Identity fidelity is evaluated based on the similarity of detected faces between the reference image and generated images. Semantic consistency is quantified using CLIP text-image consistency (CLIP-T \cite{radford2021learning}), which compares the text prompt with the corresponding generated images. More definition of metrics are detailed in Sec. \ref{sup-metrics} of the supplementary materials.

\subsection{Comparison to Previous Methods}
\label{sec-comparion}
\noindent\textbf{Raw Photo Generation.}
We benchmark our Infinite-ID against established identity-preserving personalization approaches. The qualitative outcomes are illustrated in Fig. \ref{image-realphoto-comparison}, while quantitative results are provided in Table \ref{tab-comparison} and visually represented in Fig. \ref{image-scatter}. Notably, all methods are tuning-free, necessitating no test-time adjustments. FastComposer \cite{xiao2023fastcomposer} exhibits challenges in maintaining identity fidelity, often presenting undesired artifacts in synthesized images. While IP-Adapter \cite{ye2023ip} and IP-Adapter-Face \cite{ye2023ip} demonstrates relatively fewer artifacts, its semantic consistency fall short. This phenomenon arises from the direct fusion of identity information with semantic details within the U-net model, leading to a compromise in semantic consistency. In contrast, PhotoMaker \cite{li2023photomaker} exhibits commendable semantic consistency but falls short in preserving identity fidelity. Leveraging our ID-semantics decoupling paradigm, our method excels in preserving identity fidelity. Furthermore, our mixed attention mechanism effectively integrate the semantic information into the denoising process, positioning our method favorably against existing techniques.

\begin{figure}[!tbp]
    \centering
    \includegraphics[scale=0.54]{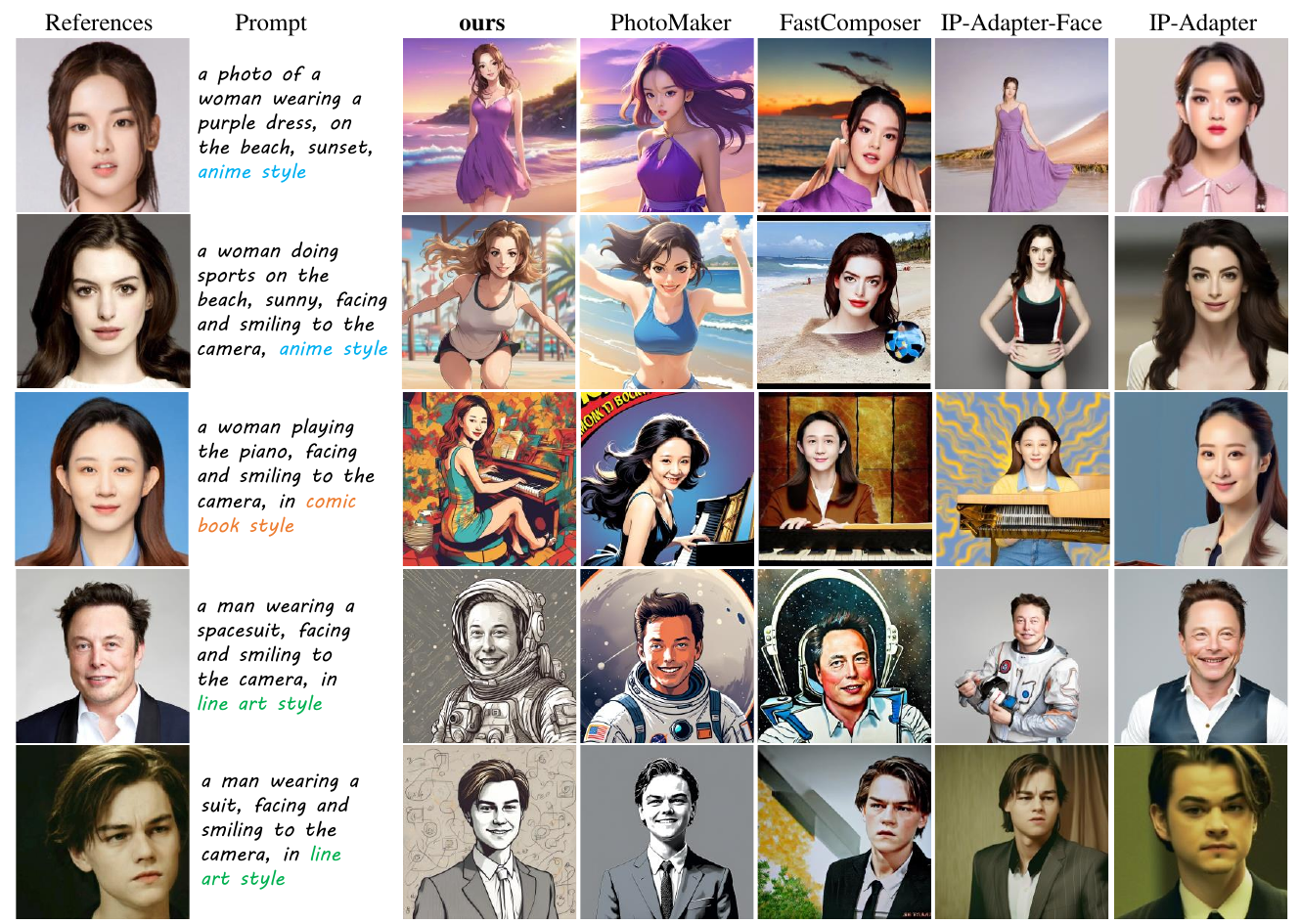}
    \caption{\textbf{Qualitative comparison on style image generation.} The results demonstrate that our method maintains strong identity fidelity, high-quality semantic consistency, and precise stylization using only a single reference image.}
    \label{image-stylephoto-comparison}
\end{figure}


\noindent\textbf{Style Image Generation.}
We demonstrate the results of the stylization results in Fig. \ref{image-stylephoto-comparison}, which compare our method with state-of-the-art tuning-free identity-preserving personalization methods. The style of the synthesis images include anime style, comic book style, and line art style. According to the stylization results, the IP-Adapter \cite{ye2023ip} and IP-Adapter-Face fails to depict the desired style in the prompt and the style of generation results always obey the tone in the reference image. The training pipeline of the IP-Adapter and IP-Adapter-Face entangles the text embedding and image embedding which leads to the distortion of the text embeddings space. FastComposer\cite{xiao2023fastcomposer} also fails in stylization generation and shows undesired artifacts. PhotoMaker\cite{li2023photomaker} achieves a better semantic consistency and stylization, but the identity fidelity is still unsatisfactory. In contrast, our method achieves high identity fidelity, appealing semantic consistency, and precise stylization meantime.

\begin{figure}[!tbp]
    \centering
    \includegraphics[scale=0.5]{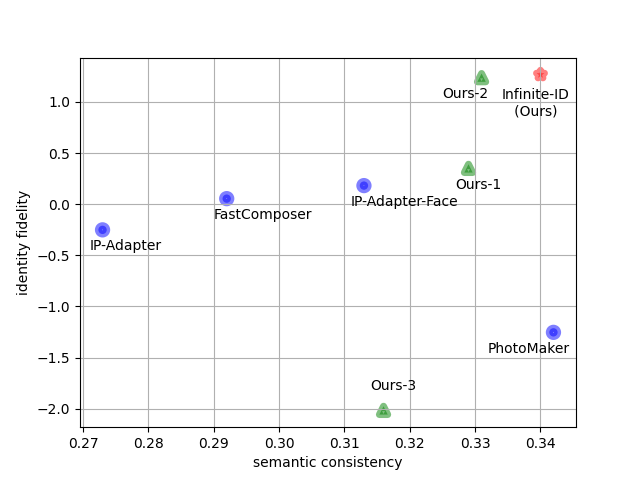}
    \caption{\textbf{Visualization of the quantitative comparison.} 
Identity fidelity, represented by the average of CLIP-I and $M_\text{FaceNet}$ scores, both normalized through the z-score algorithm, indicates how accurately the generated image preserves the identity. Meanwhile, semantic consistency, measured by the CLIP-T score, assesses the coherence between the generated image and the provided text prompt. Higher scores indicate better identity fidelity and semantic consistency. The compared methods including IP-Adapter, IP-Adapter-Face\cite{ye2023ip}, FastComposer\cite{xiao2023fastcomposer}, PhotoMaker\cite{li2023photomaker}, and ablation versions of our method including w/o identity-enhanced training (Ours-1), w/o mixed attention (Ours-2) and mixed attention $\Rightarrow$ mutual attention (Ours-3).}
    \label{image-scatter}
\end{figure}

 \begin{table}[!tbp]
    \centering
    \caption{\textbf{Quantitative comparison.} The evaluation metrics encompass CLIP-T, CLIP-I, and $M_\text{FaceNet}$. Our approach outperforms other methods in terms of identity fidelity while simultaneously achieving satisfactory semantic consistency. The best result is shown in \textbf{bold}, and the second best is \underline{underlined}. Additionally, the quantitative comparison of ablation studies have been shown in the gray part.}
    \begin{tabular}{c|c|c|c}
    \hline
                                        &CLIP-T $\uparrow$       &CLIP-I $\uparrow$         &$M_\text{FaceNet}$ $\uparrow$\\
    \hline
FastComposer\cite{xiao2023fastcomposer} &0.292                   &0.887                     &\underline{0.556}\\
    \hline
IP-Adapter\cite{ye2023ip}          &0.274                   &0.905         &0.474\\
\hline
IP-Adapter-Face\cite{ye2023ip}      &0.313                  &\textbf{0.919}                      &0.513\\
\hline
PhotoMaker\cite{li2023photomaker}       &\textbf{0.343}          &0.814                     &0.502\\
\hline
\cellcolor[gray]{0.85} Ours                                    &\cellcolor[gray]{0.85}\underline{0.340}       &\cellcolor[gray]{0.85}\underline{0.913}            &\cellcolor[gray]{0.85}\textbf{0.689}\\
\hline
\cellcolor[gray]{0.85}Ours (w/o identity-enhanced training)        &\cellcolor[gray]{0.85}0.329                   &\cellcolor[gray]{0.85}0.891                    &\cellcolor[gray]{0.85}0.593\\
    \hline
\cellcolor[gray]{0.85}Ours (w/o Mixed Attention)                         &\cellcolor[gray]{0.85}0.331                   &\cellcolor[gray]{0.85}0.905                    &\cellcolor[gray]{0.85}{0.700}\\
\hline
\cellcolor[gray]{0.85}Ours (Mixed Attention $\Rightarrow$ Mutual Attention)     &\cellcolor[gray]{0.85}0.316                   &\cellcolor[gray]{0.85}0.808                    &\cellcolor[gray]{0.85}0.398\\
\hline
    \end{tabular}
    \label{tab-comparison}
\end{table}


\subsection{Ablation Study}
\label{sec:ablation}
In this section, we begin by conducting ablation studies to assess the influence of our identity-enhanced training and mixed attention mechanism. Furthermore, we conduct style image generation to evaluate the effectiveness of our AdaIN-mean operation in regulating the style of the generated images. More ablation studies are demonstrated in Sec. \ref{sup-abla} of the supplementary materials.



\begin{figure}[!tbp]
    \centering
    \includegraphics[scale=0.67]{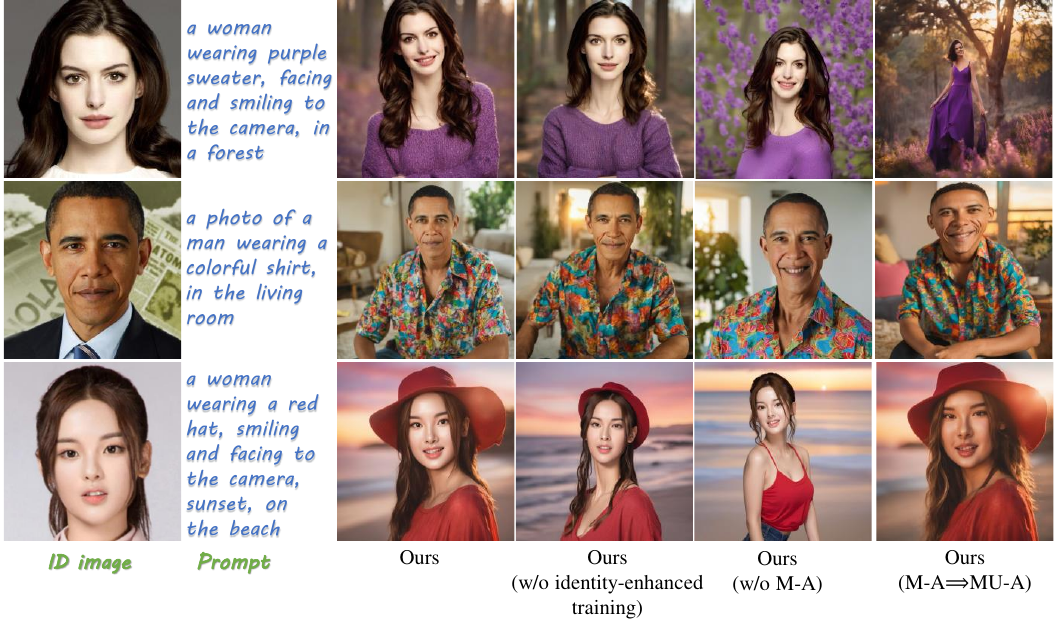}
    \caption{\textbf{Ablation study of our identity-enhanced training and mixed attention ($\text{M-A}$).} It is evident that identity-enhanced training significantly improves the identity fidelity, and mixed-attention mechanism enhances semantic consistency compared to mutual attention ($\text{MU-A}$) approach \cite{cao2023masactrl}.}
    \label{image-ablation}
\end{figure}

\noindent\textbf{Ablation of identity-enhanced training.}
In Fig. \ref{image-ablation} and Table \ref{tab-comparison}, we compare our method with "Ours (w/o identity-enhanced training)", which is implemented using an identity-semantics entangled training strategy. This strategy utilizes text-image pairs during training and activates cross-attention modules for text embeddings within the original U-Net model. It is noteworthy that both methods share the same inference processing. The qualitative comparison demonstrates that our identity-enhanced training notably enhances identity fidelity.


\noindent\textbf{Ablation of mixed attention mechanism.}
To assess the effectiveness of our proposed mixed attention (M-A) mechanism, we compare our method with "Ours (w/o M-A)" and "Ours (M-A $\Rightarrow$ MU-A)" in Fig. \ref{image-ablation} and Table \ref{tab-comparison}. Mutual self-attention (MU-A) \cite{cao2023masactrl} converts the existing self-attention into ‘cross-attention’ for consistent image editing, where the crossing operation happens in the self-attentions of two related diffusion processes. The results show that our mixed attention mechanism demonstrates superior ability to improve semantic consistency while maintaining identity fidelity.

\begin{figure}[!tbp]
    \centering
    \includegraphics[scale=0.51]{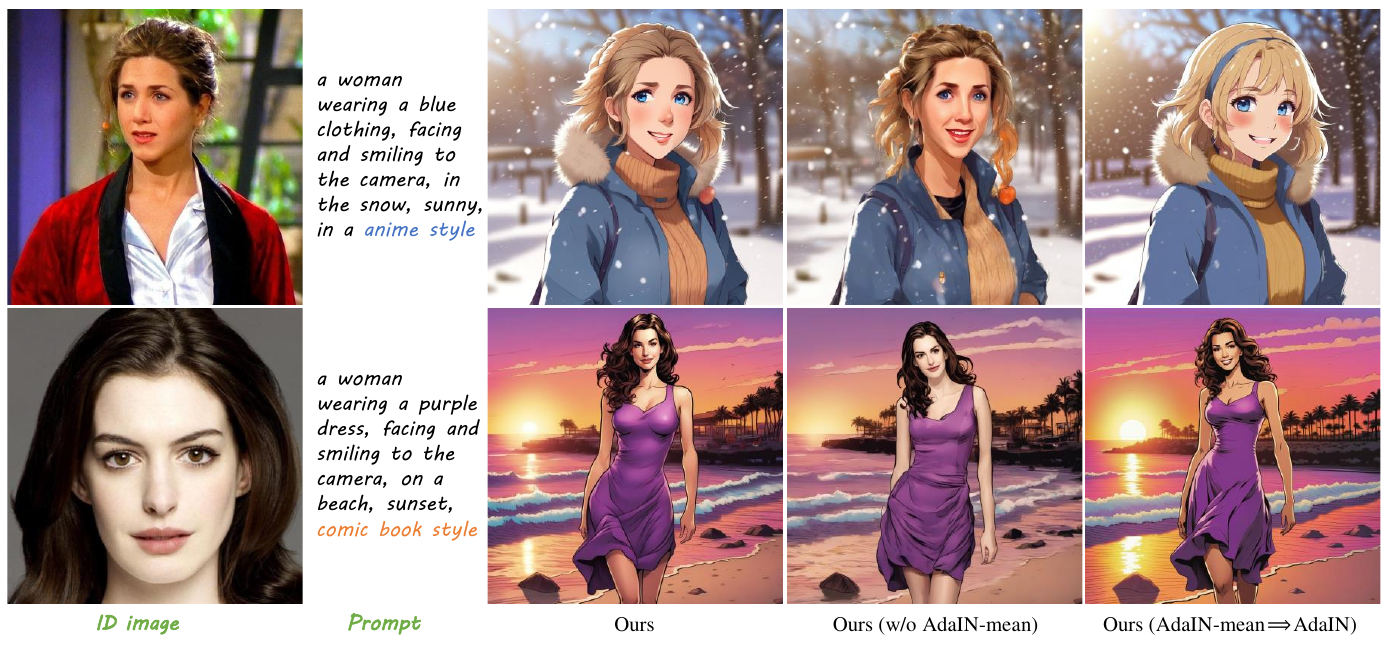}
    \caption{\textbf{Ablation study of our AdaIN-mean operation.} The results show that AdaIN-mean play a crucial role in style image generation. Compare to AdaIN\cite{hertz2023style} module, our AdaIN-mean helps to achieve a higher identity fidelity.}
    \label{image-ablation-adain}
\end{figure}

\noindent\textbf{Ablation of AdaIN-mean operation.} 
To assess the effectiveness of our proposed AdaIN-mean operation, we compare our Infinite-ID model with variations, namely Ours (w/o AdaIN-mean) and Ours (AdaIN-mean $\Rightarrow$ AdaIN), as depicted in Fig. \ref{image-ablation-adain}. The results reveal that: i) Ours (w/o AdaIN-mean) exhibits superior ID fidelity compared to Infinite-ID but fails to achieve style consistency with the text prompt; ii) Both our AdaIN-mean and AdaIN modules successfully achieve style consistency with the text prompt, yet AdaIN-mean maintains better ID fidelity than AdaIN. In conclusion, our proposed AdaIN-mean operation facilitates precise stylization while concurrently preserving ID fidelity.



\section{Conclusion and limitations.}


In this paper, we introduce Infinite-ID, an innovative identity-preserved personalization method designed to meet the requirement of identity (ID) fidelity and semantic consistency of the text prompt, all achievable with just one reference image and completed within seconds. Infinite-ID comprises three key components: the identity-enhanced training, mixed attention mechanism, and adaptive mean normalization (AdaIN-mean). Through extensive experimentation, our results illustrate that Infinite-ID outperforms baseline methods, delivering strong ID fidelity, superior generation quality, and precise semantic consistency in both raw photo generation and style image generation tasks. However, it's important to note that our method lacks multi-object personalization capability. Moreover, artifacts may occur when the human face occupies only a small portion of the entire image, attributed to limitations inherent in the original diffusion model.




%
%
\bibliographystyle{splncs04}
\bibliography{main}

\newpage

\appendix
\section{Supplementary Materials}

\subsection{Evaluation metrics}
\label{sup-metrics}


\noindent\textbf{Evaluation of face similarity.} 
To assess face similarity, we utilize the face alignment module $\text{FA}(\cdot)$, the face recognition backbone $\text{E}_{\text{face}}(\cdot)$, and the CLIP image encoder $\text{E}_{\text{clip}}(\cdot)$ to compute the metrics $M_{\text{FaceNet}}$ and $\text{CLIP}\text{-}{\text{I}}$. Specifically, for each generated image $\text{I}_{\text{gen}}$ and its corresponding identity image $\text{I}_{\text{id}}$, we first employ the $\text{FA}(\cdot)$ module to detect the face. Subsequently, we calculate the pairwise identity similarity using $\text{E}_{\text{face}}(\cdot)$ and $\text{E}_{\text{clip}}(\cdot)$, respectively:
\begin{equation}
\begin{aligned}
    M_{\text{FaceNet}} = cos(\text{E}_{\text{face}}(\text{FA}(\text{I}_{\text{gen}})), \text{E}_{\text{face}}(\text{FA}(\text{I}_{\text{id}}))),\\
    \text{CLIP}\text{-}{\text{I}} = cos(\text{E}_{\text{clip}}(\text{FA}(\text{I}_{\text{gen}})), \text{E}_{\text{clip}}(\text{FA}(\text{I}_{\text{id}}))),
\end{aligned}
\end{equation}
where $cos(\cdot, \cdot)$ is the cosine similarity function. Furthermore, in order to illustrate the identity fidelity depicted in Figure 6 of the main paper, we integrate both $M_{\text{FaceNet}}$ and $\text{CLIP}\text{-}{\text{I}}$ by utilizing z-score normalization:
\begin{equation}
    mean(\text{z-score}(M_{\text{FaceNet}}), \text{z-score}(\text{CLIP-I})),
\end{equation}
where $\text{z-score}(x)=(x-\mu)/\sigma$, $\mu$ and $\sigma$ are the average and standard deviation of the $x$, respectively.

\noindent\textbf{Definition of semantic consistency.} 
We adopt the CLIP-T metric to assess semantic consistency. Specifically, for a generated image $I_{\text{gen}}$ paired with its corresponding prompt $P$, we compute the CLIP-T metric utilizing both the CLIP image encoder $E_{\text{clip}}$ and the CLIP text encoder $E_{\text{text}}$:
\begin{equation}
    \text{CLIP-T}=cos(E_\text{clip}(I_{gen}),E_\text{text}(P)),
\end{equation}
where the $cos(\cdot, \cdot)$ is the cosine similarity function.

\subsection{More Results on Ablation Study}
\label{sup-abla}

\begin{figure}[!tbp]
    \centering
    \includegraphics[scale=0.6]{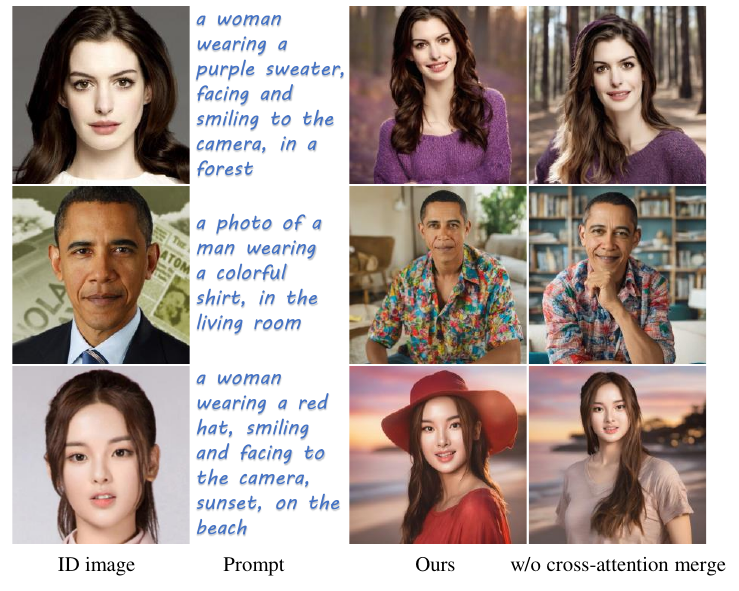}
    \caption{\textbf{Quantitative ablation of cross-attention merge.} It is obvious that the cross-attention merge helps to improve the semantic consistency.}
    \label{sup-abla-cross}
\end{figure}

\noindent\textbf{Ablation Study of Cross-attention Merge.}
We conduct ablation experiments on the cross-attention merge to evaluate its effectiveness. As depicted in Fig. \ref{sup-abla-cross} and Table \ref{sup-tab-ablation-1}, the incorporation of cross-attention merge demonstrates improvement in semantic consistency.

\begin{figure}[!htbp]
    \centering
    \includegraphics[scale=0.36]{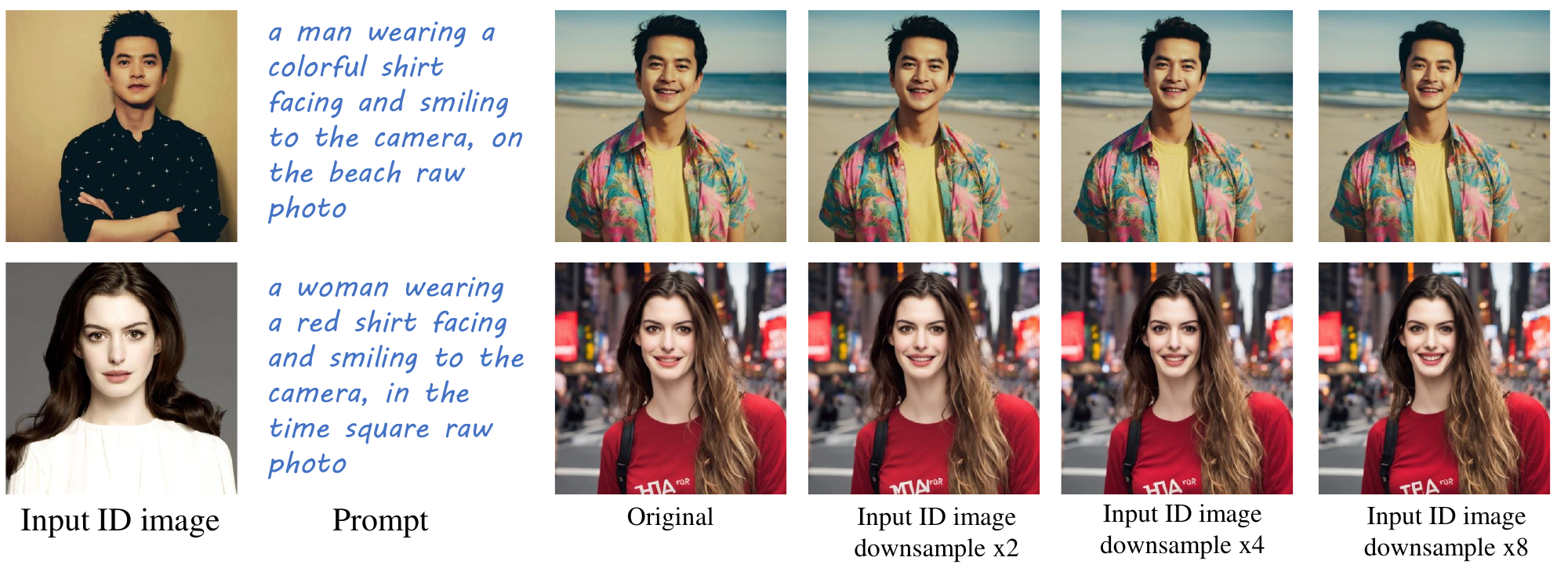}
    \caption{\textbf{Ablation study of input ID images' resolution}. The identity fidelity slightly drops along with the lower image resolution and the semantic consistency is stable for all the resolution. Our method is robust to the resolution of input ID image.}
    \label{sup-multi-res}
\end{figure}

\noindent\textbf{Ablation Study of Input ID Images' Resolution.}
We perform an ablation study on the resolution of input ID images to assess the robustness of our method. Specifically, we utilize images with varying resolutions while maintaining the same text prompt for personalization. As illustrated in Fig. \ref{sup-multi-res}, the identity fidelity exhibits only a marginal decrease with decreasing image resolution, while semantic consistency remains stable across all resolutions. In conclusion, our method demonstrates robustness to changes in input image resolution.

 \begin{table}[!tbp]
    \centering
    \caption{\textbf{Quantitative ablation of cross-attention merge.} The metrics includes CLIP-T (higher is better) measuring the semantic consistency, CLIP-I (higher is better) and $M_\text{FaceNet}$ (higher is better) which are both reflect the identity fidelity. The best result is shown in \textbf{bold}.}
    \begin{tabular}{c|c|c|c}
    \hline
                                                 &CLIP-T$\uparrow$        &CLIP-I$\uparrow$         &$M_\text{FaceNet}$$\uparrow$\\
    \hline
Ours w/o Cross-attention merge                   &0.335       &0.910        &0.681\\
\hline
Ours                                             &\textbf{0.340}          &\textbf{0.913}           &\textbf{0.689}\\
\hline
    \end{tabular}
    \label{sup-tab-ablation-1}
\end{table}

\subsection{Identity mixing}
\label{sup-idmixed}
\begin{figure}[!htbp]
    \centering
    \includegraphics[scale=0.38]{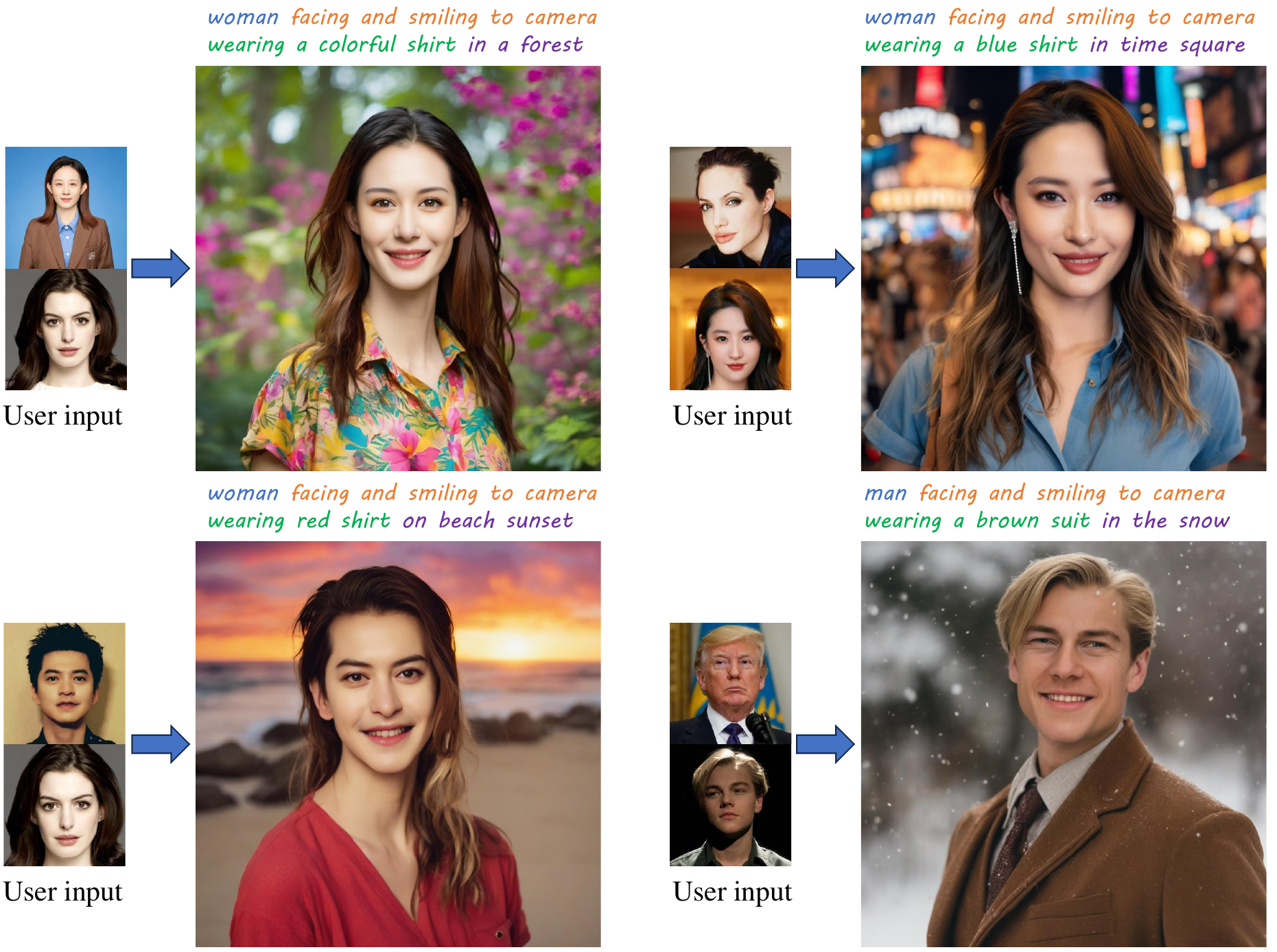}
    \caption{\textbf{Identity mixing.} When receiving multiple input ID images from different individuals, our method can mix these identities by stacking all the identity embeddings.}
    \label{sup-multi-diff}
\end{figure}

\begin{figure}[!htbp]
    \centering
    \includegraphics[scale=0.27]{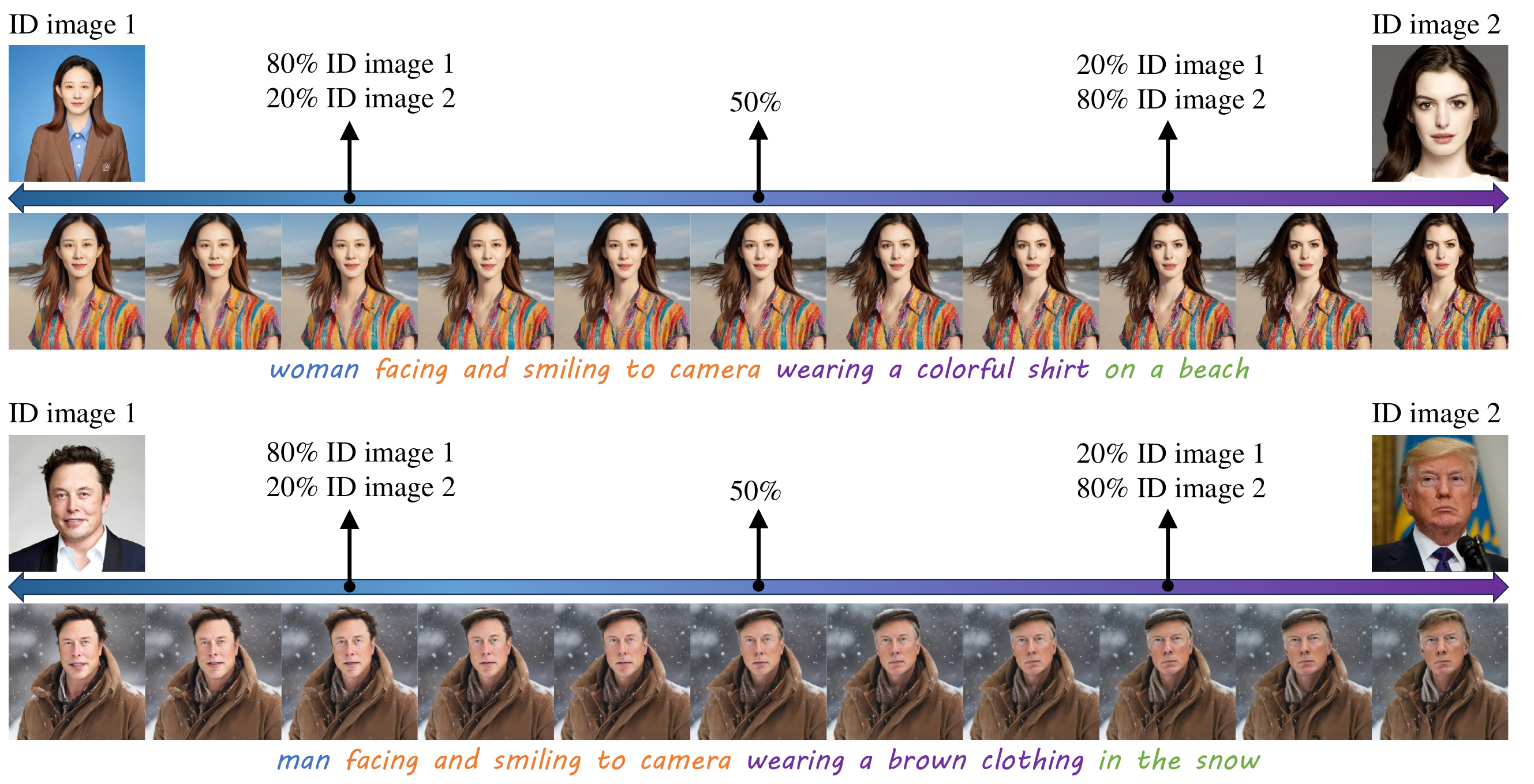}
    \caption{\textbf{Linear interpolation of different identities.}}
    \label{sup-multi-diff-2}
\end{figure}

Upon receiving multiple images from distinct individuals, we stack all the identity embeddings to merge corresponding identities, as depicted in Fig. \ref{sup-multi-diff}. The generated image can well retain the characteristics of different IDs, which releases possibilities for more applications. Additionally, by adjusting the interpolation of the identity embeddings, we can regulate the similarity between the generated identity and different input identities, as demonstrated in Fig. \ref{sup-multi-diff-2}.

\subsection{More Qualitative Results of Raw Photo Generation}
\label{sup-rawphoto}
\begin{figure}[!htbp]
    \centering
    \includegraphics[scale=0.37]{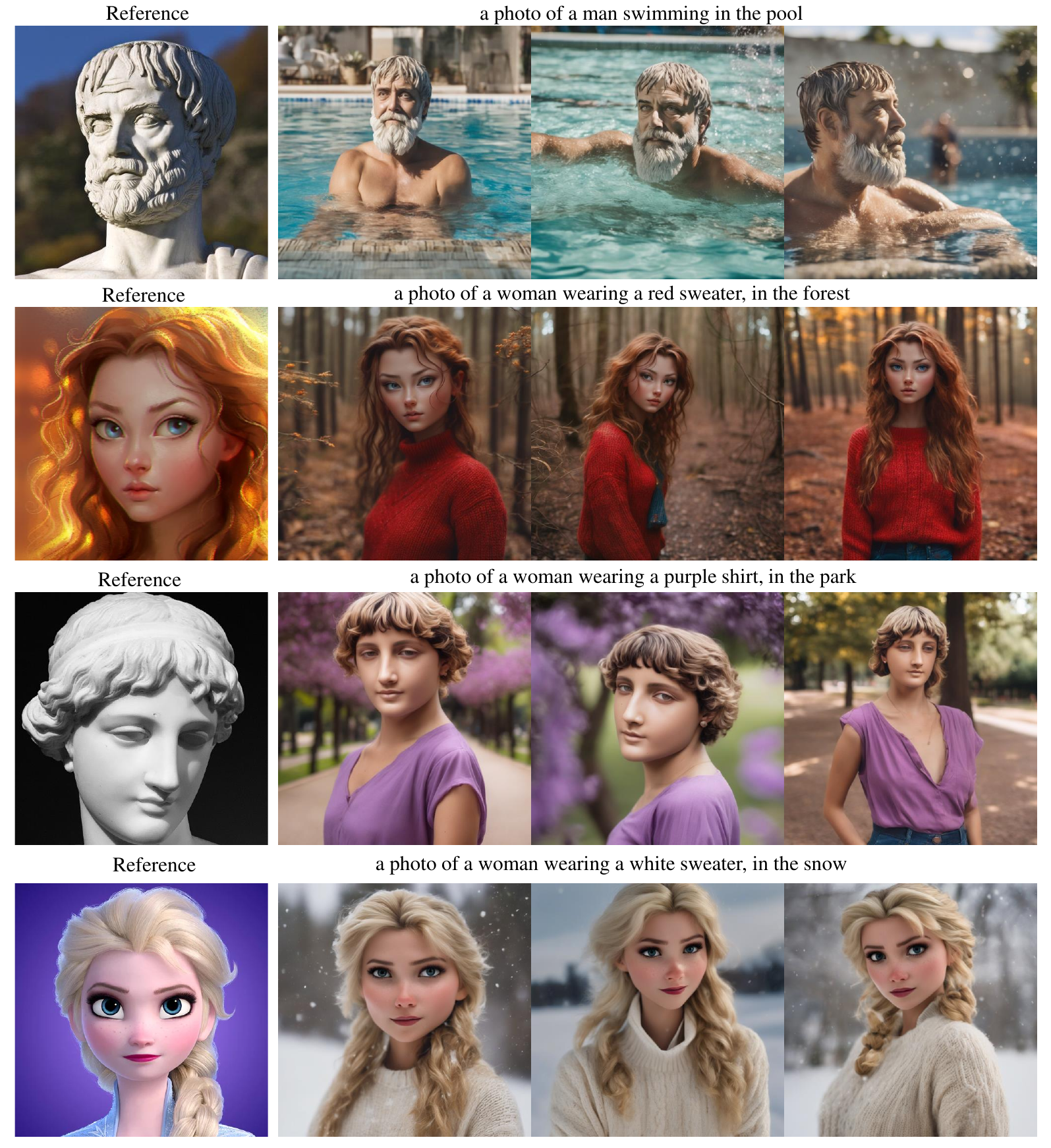}
    \caption{\textbf{Applications on artworks to raw photo.}}
    \label{sup-raw-photo-1}
\end{figure}

\begin{figure}[!htbp]
\vspace{-1cm}
    \centering
    \includegraphics[scale=0.37]{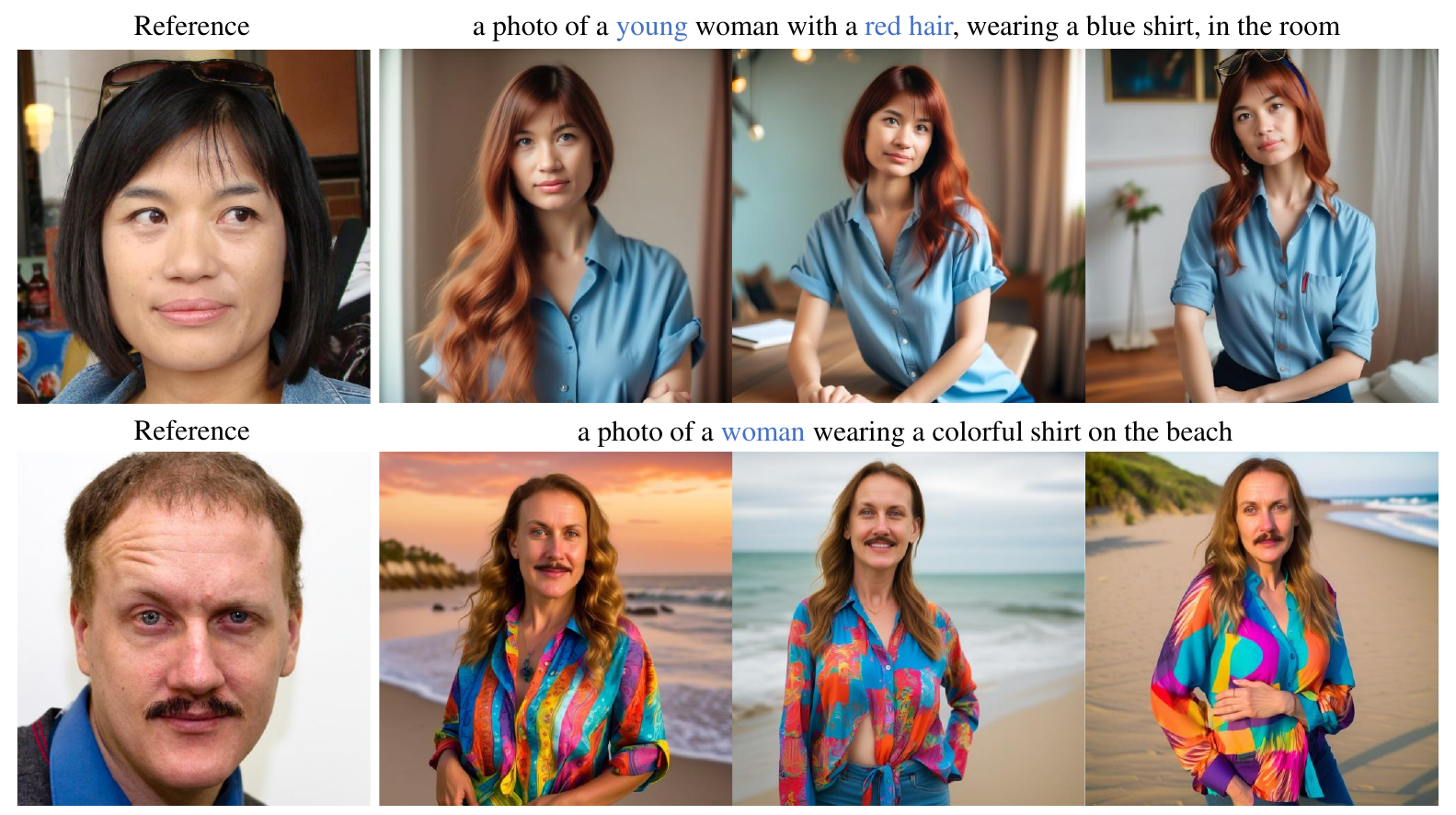}
    \caption{\textbf{Applications on attribute change.}}
    \label{sup-raw-photo-2}
\end{figure}

\begin{figure}[!htbp]
    \centering
    \includegraphics[scale=0.4]{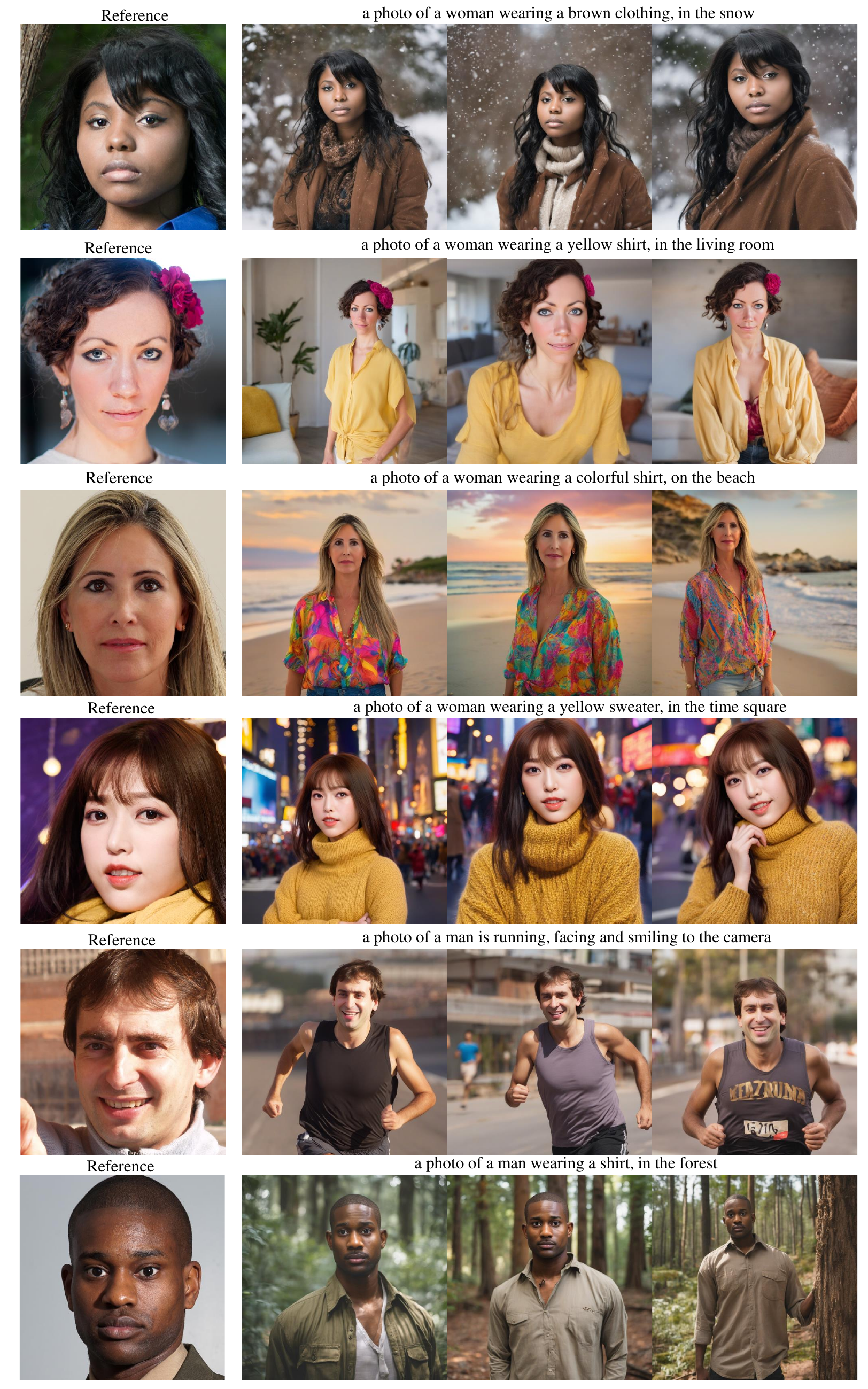}
    \caption{\textbf{Raw photo generation.} These identities are ordinary people sampled from FFHQ dataset, including various races, skin colors, male and female.}
    \label{sup-raw-photo-3}
\end{figure}

\begin{figure}[!htbp]
    \centering
    \includegraphics[scale=0.4]{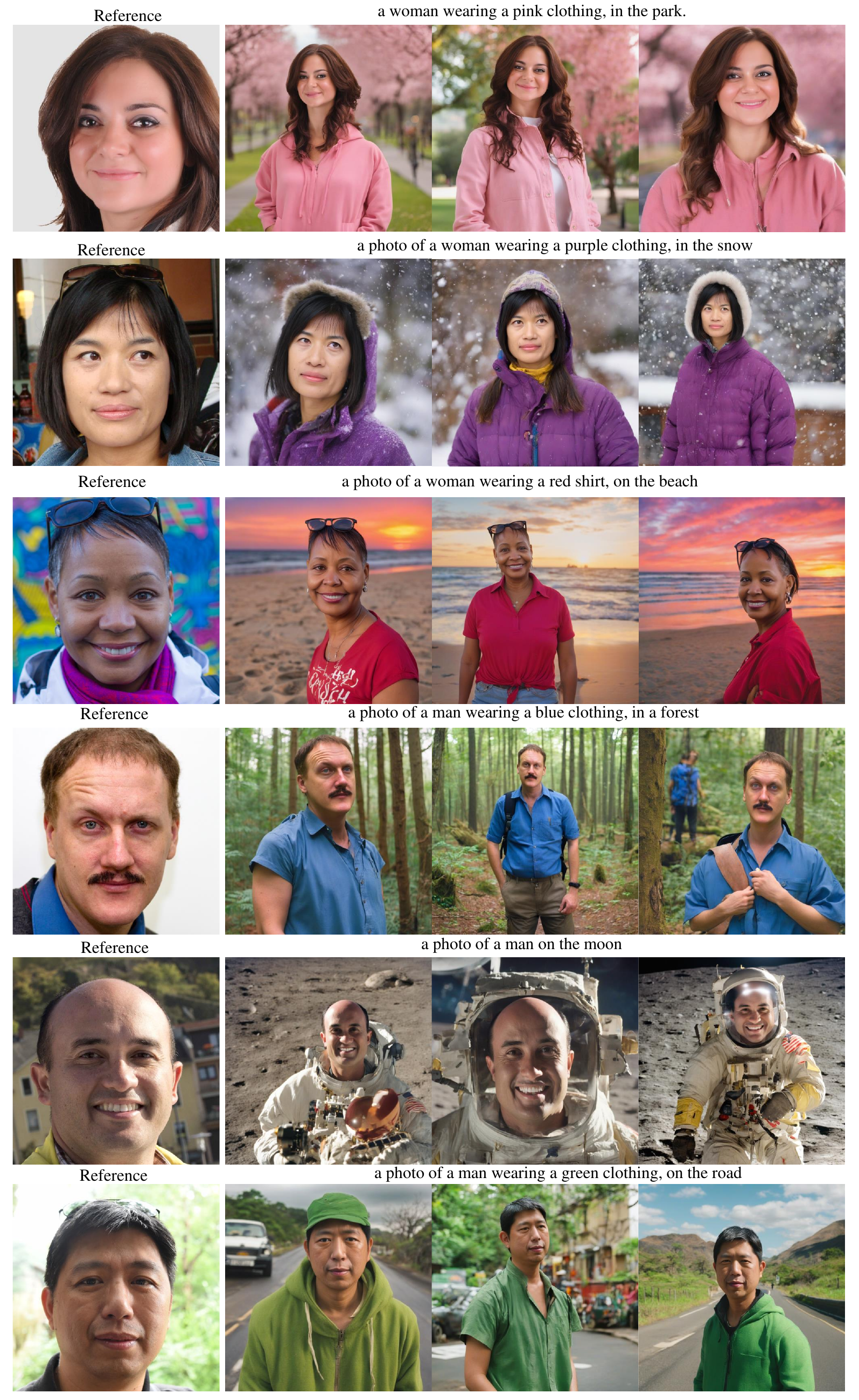}
    \caption{\textbf{Raw photo generation.} These identities are ordinary people sampled from FFHQ dataset, including various races, skin colors, male and female.}
    \label{sup-raw-photo-4}
\end{figure}

Fig. \ref{sup-raw-photo-1} demonstrates the ability of our method to extract identity information from artworks while preserving identity for personalization purposes. Additionally, Fig. \ref{sup-raw-photo-2} illustrates the capability of our method to alter attributes of the extracted identities for raw photo generation. Additional visual samples for raw photo generation are provided in Fig. \ref{sup-raw-photo-3} and Fig. \ref{sup-raw-photo-4}, showcasing identities of ordinary individuals sampled from the FFHQ dataset, spanning diverse races, skin tones, and genders.

\subsection{More Qualitative Results of Style Photo Generation}
\label{sup-stylephoto}
\begin{figure}[!htbp]
    \centering
    \includegraphics[scale=0.4]{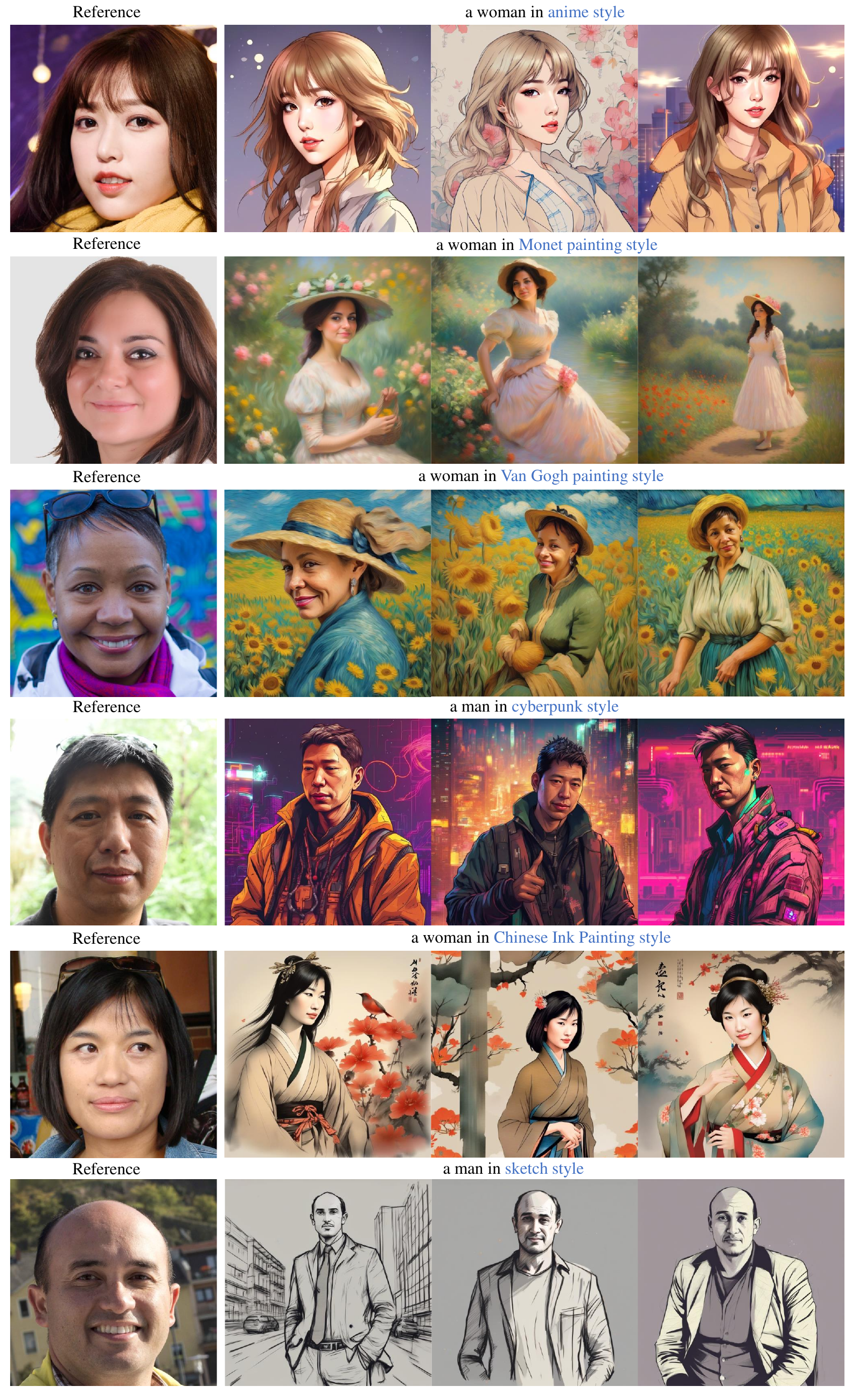}
    \caption{\textbf{More visual examples for stylization.} These identities are ordinary people sampled from FFHQ dataset, including various races, skin colors, male and female.}
    \label{sup-style-photo-1}
\end{figure}

\begin{figure}[!htbp]
    \centering
    \includegraphics[scale=0.4]{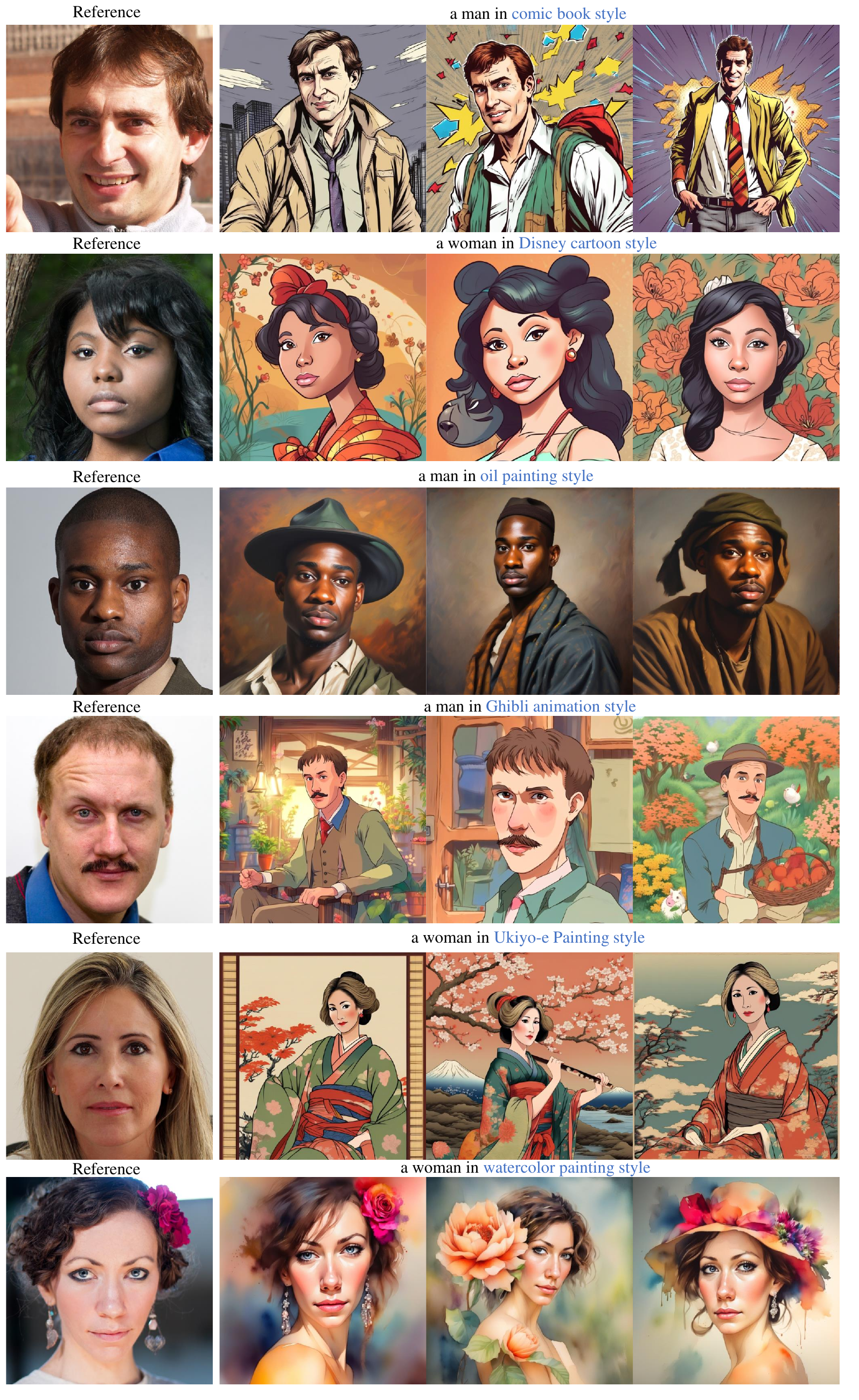}
    \caption{\textbf{More visual examples for stylization.} These identities are ordinary people sampled from FFHQ dataset, including various races, skin colors, male and female.}
    \label{sup-style-photo-2}
\end{figure}

Fig. \ref{sup-style-photo-1} and Fig. \ref{sup-style-photo-2} display the results of style photo generation. The identity samples consist of ordinary individuals randomly selected from the FFHQ dataset. A total of 12 stylization styles are employed, affirming the generalizability of our method.

\end{document}